\documentclass{article}
\usepackage{ijcai25}

\usepackage{times}
\usepackage{soul}

\usepackage{url}
\usepackage[usenames,svgnames]{xcolor}
\usepackage[colorlinks,
citecolor=NavyBlue,
linkcolor=NavyBlue,
urlcolor=NavyBlue]{hyperref}

\usepackage[utf8]{inputenc}
\usepackage[small]{caption}
\usepackage{graphicx}
\usepackage{amsmath}
\usepackage{amsthm}
\usepackage{booktabs}
\usepackage{algorithm}
\usepackage{algorithmic}
\usepackage[switch]{lineno}
\usepackage{enumitem}

\usepackage{amsfonts}  

\usepackage{mydef}

\usepackage{booktabs}
\usepackage{tabularx}
\usepackage{framed}
\usepackage{bbding}
\usepackage{subfloat}
\usepackage{adjustbox}
\usepackage{multirow}
\usepackage{pifont}
\usepackage{natbib}

\usepackage{makecell}

\usepackage[capitalize]{cleveref}
\crefname{figure}{Figure}{Figures}
\crefname{table}{Table}{Tables}
\crefformat{equation}{Eq.~(#2#1#3)}

\newcommand{\annotator}[2]{\csdef{#1}##1{{\color{#2} [\textbf{\MakeUppercase #1}: ##1]}}}
\annotator{leon}{blue}

\newcommand{\paratitle}[1]{\vspace{0.8ex}\noindent \textbf{#1}}

\title{Diffusion Models for Molecules: A Survey of Methods and Tasks}

\author{
Liang Wang$^{1,2}$
\and
Chao Song$^3$\and
Zhiyuan Liu$^{4}$\and
Yu Rong$^{5}$\and \\
Qiang Liu$^{1,2}$\and 
Shu Wu$^{1,2}$\And
Liang Wang$^{1,2}$\\
\affiliations
$^1$NLPR, MAIS, Institute of Automation, Chinese Academy of Sciences\\
$^2$School of Artificial Intelligence, University of Chinese Academy of Sciences\\
$^3$Northwestern Polytechnical University\\
$^4$National University of Singapore\\
$^5$Alibaba DAMO Academy\\
\emails
liang.wang@cripac.ia.ac.cn,
csong@mail.nwpu.edu.cn,
acharkq@gmail.com,\\
yu.rong@hotmail.com,
\{qiang.liu, shu.wu, wangliang\}@nlpr.ia.ac.cn
}

\begin{document}

\maketitle

\begin{abstract}
    Generative tasks about molecules, including but not limited to molecule generation, are crucial for drug discovery and material design, and have consistently attracted significant attention. In recent years, diffusion models have emerged as an impressive class of deep generative models, sparking extensive research and leading to numerous studies on their application to molecular generative tasks. Despite the proliferation of related work, there remains a notable lack of up-to-date and systematic surveys in this area. Particularly, due to the diversity of diffusion model formulations, molecular data modalities, and generative task types, the research landscape is challenging to navigate, hindering understanding and limiting the area's growth. To address this, this paper conducts a comprehensive survey of diffusion model-based molecular generative methods. We systematically review the research from the perspectives of methodological formulations, data modalities, and task types, offering a novel taxonomy. This survey aims to facilitate understanding and further flourishing development in this area.
    The relevant papers are summarized at: \url{https://github.com/AzureLeon1/awesome-molecular-diffusion-models}.
\end{abstract}

\section{Introduction}

\begin{figure}[t]
  \centering
  \includegraphics[width=0.4\textwidth]{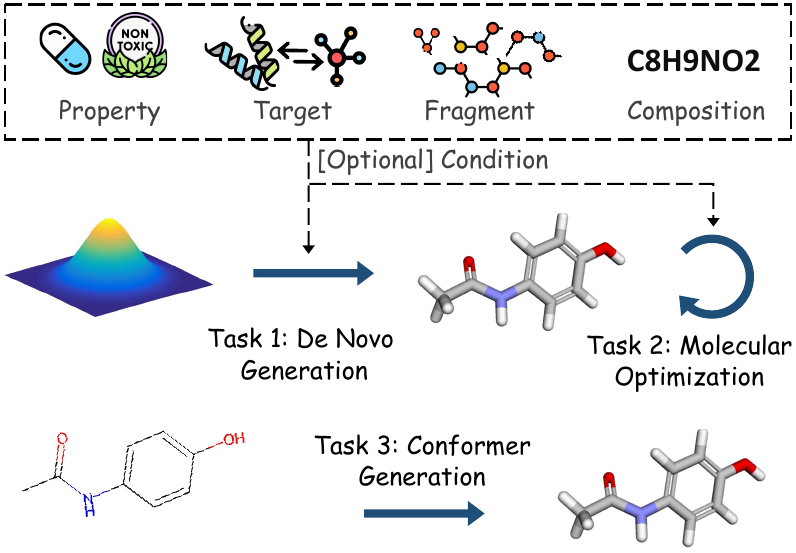}
  \caption{Illustration of molecular generative tasks. \textit{De novo generation} designs molecules from scratch. \textit{Molecular optimization} refines existing molecules to enhance desired properties while maintaining structure similarity. \textit{Conformer generation} generates 3D geometries of a molecule to represent its possible spatial arrangements.}
  \label{fig:overview}
  \vspace{-6pt}
\end{figure}

Molecular generative tasks (\cref{fig:overview}) are critical for drug discovery and material design~\citep{InverseDesignSurvey,MolGenSurvey1,MolGenSurvey2}. The ability to generate novel molecules with specific desired properties can significantly expedite the development of new pharmaceuticals and advanced materials, thereby addressing pressing challenges in healthcare and technology. However, traditional methods for these tasks are often labor intensive and time-consuming. 

Deep generative models, such as Variational Autoencoders (VAEs)~\citep{VAE}, Generative Adversarial Networks (GANs)~\citep{GAN}, Autoregressive models (ARs), and Normalizing Flows (NFs)~\citep{NormalizingFlow}, have opened new avenues for automating molecular generative tasks, gaining prominence for their ability to explore vast chemical spaces efficiently. These models enhance the speed and accuracy of molecular discovery, making them indispensable tools in modern scientific research.

Diffusion models~\citep{Diffusion,DDPM} have recently emerged as powerful generative models, showcasing remarkable performance in generating high-quality data across various domains. 
These models operate by simulating the gradual degradation of a data distribution and learning its reverse process to generate new samples~\citep{understanding}. 
Originally introduced for visual domains~\citep{DiffusionVisionSruvey}, these models excel at capturing data distributions through iterative processes. Their success in visual domains has inspired researchers to explore their potential for molecular generative tasks. By effectively modeling intricate molecular structures and properties, diffusion models have become central to molecular design. This has sparked a surge in research adapting diffusion models for molecular applications, highlighting their transformative potential in this area~\citep{GDSS,EDM}.

Despite the rapid advancements and the proliferation of research on diffusion model-based molecular design, the area faces significant challenges. \textit{The diversity in diffusion model formulations, molecular data modalities, and generative task types has resulted in a fragmented research landscape.} This diversity makes it difficult for researchers to navigate existing studies, hindering a comprehensive understanding of the area's progress and potential. The lack of a systematic and up-to-date survey exacerbates this issue, as researchers struggle to keep pace with the latest developments. This fragmentation not only limits the accessibility of existing advances, but also constrains the area's growth and innovation.

To address these challenges, this paper presents an comprehensive survey of diffusion model-based molecular generative methods. We provide a comprehensive and systematic review of the area, elucidating the design space of existing works
according to method formulations, data modalities, and task types, as shown in \cref{fig:taxonomy} and \cref{tab:summarizations}. By offering this novel taxonomy, we clarify the research landscape and facilitate easier navigation for researchers. 

This survey seeks to bridge the gap between diverse studies, promoting a more cohesive understanding of the area and supporting its further development. By highlighting key advancements and identifying emerging opportunities for future research, we hope to contribute to the area's ongoing evolution and encourage future innovations.

This survey makes several key contributions to the area of molecular generative tasks using diffusion models:
\begin{itemize}[leftmargin=*]
    \item We provide an up-to-date and systematic overview of the current state of research, addressing the need for a comprehensive understanding of the area.
    \item We introduce a novel taxonomy that categorizes research efforts based on method formulations, data modalities, and task types, offering a structured framework.
    \item By identifying opportunities in the existing literature, we pave the way for future directions, encouraging further innovation in diffusion model-based molecular design.
\end{itemize}
\section{Formulations of Diffusion Models}\label{sec:diffusion}

\begin{figure*}[t]
    \centering
    \resizebox{0.91\linewidth}{!}{
    \includegraphics{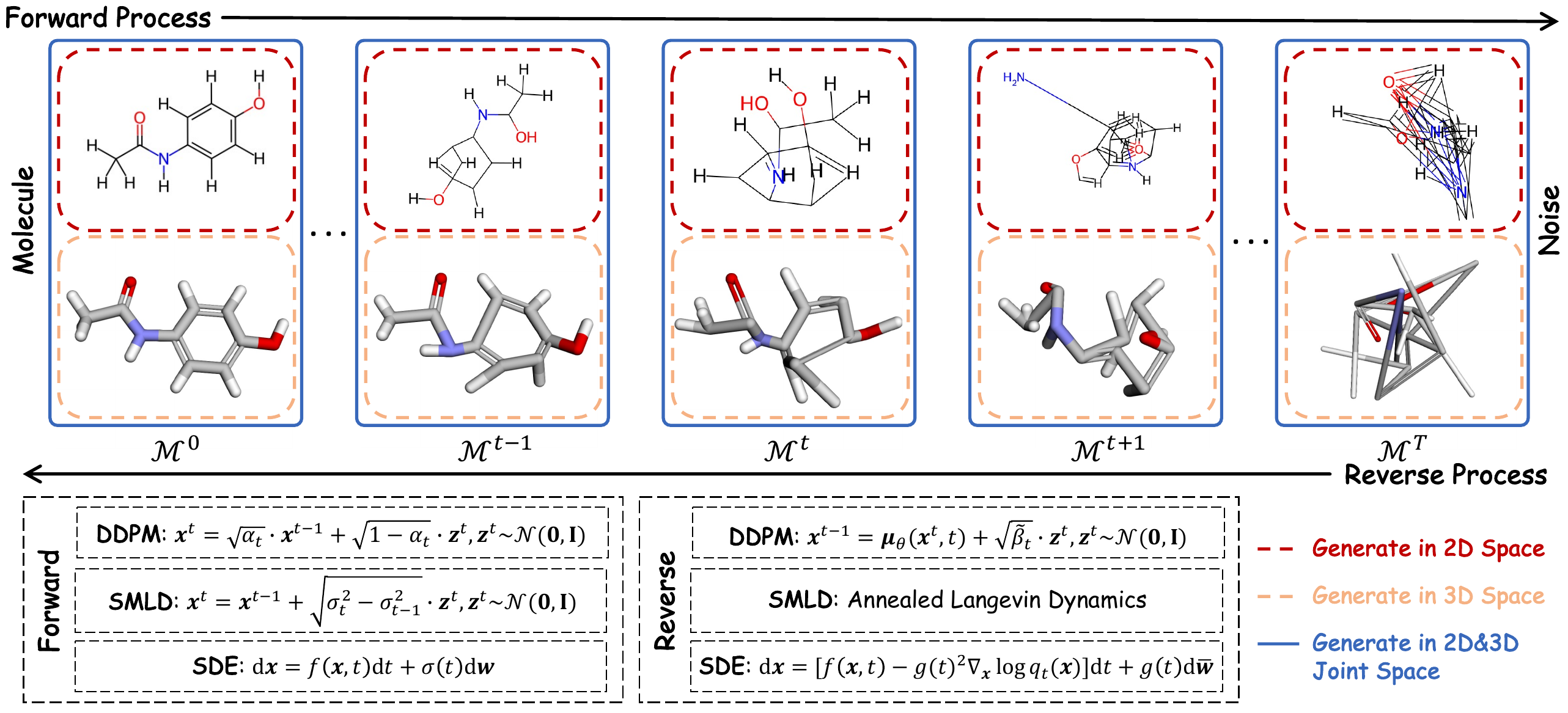}}
    \caption{Illustration of molecular diffusion models, showcasing the forward and reverse processes. The three primary formulations—DDPM, SMLD, and SDE—are presented. Molecules can be generated in 2D space, 3D space, or jointly in 2D and 3D spaces.} 
    \label{fig:diffusion}
    \vspace{-6pt}
\end{figure*}

This fundamental framework of diffusion models comprises two parts: the \textit{forward diffusion process} and the \textit{reverse generation process}, as shown in \cref{fig:diffusion}. In the forward process, the model progressively adds noise to real data, eventually approaching a simple prior distribution.
In the reverse process, the model learns to progressively restore the data distribution from noise.
The reverse process is typically parameterized using neural networks.

This framework can be formulated in various ways, such as the Denoising Diffusion Probabilistic Models~\citep{DDPM}, Score Matching with Langevin Dynamics~\citep{SMLD}, the generalized Stochastic Differential Equations~\citep{SDE}, and other variants.

\subsection{Denoising Diffusion Probabilistic Models (DDPMs)}

DDPM~\citep{Diffusion,DDPM} is a classical diffusion model that performs step-by-step denoising using a fixed noise schedule. It employs two Markov chains for the forward and reverse processes. 

Starting with the original noise-free data $\bx_0$, the forward process transforms it into a sequence of noisy data $\bx_1, \bx_2, \ldots, \bx_T$ using a forward transition kernel:
\begin{equation}
q(\bx_t|\bx_{t-1}) = \mathcal{N}\left(\bx_t; \sqrt{\alpha_t}\bx_{t-1}, (1-\alpha_t)\bI\right),
\end{equation}
where $\alpha_t \in (0, 1)$ for $t = 1, 2, ..., T$ are hyperparameters that define the noise ratio at each step. $\mathcal{N}(\bx; \boldsymbol{\mu}, \mathbf{\Sigma})$ denotes a Gaussian distribution with mean $\boldsymbol{\mu}$ and covariance $\mathbf{\Sigma}$. A useful property of this Gaussian transition kernel is that $\bx_t$ can be directly derived from $\bx_0$ by:
\begin{equation}
q(\bx_t|\bx_0) = \mathcal{N}\left(\bx_t; \sqrt{\bar{\alpha}_t}\bx_0, (1-\bar{\alpha}_t)\bI\right),
\label{eq:forward-property}
\end{equation}
where $\bar{\alpha}_t := \prod_{i=1}^t \alpha_i$. Thus, $\bx_t$ is given by $\bx_t = \sqrt{\bar{\alpha}_t}\bx_0 + \sqrt{1-\bar{\alpha}_t}\boldsymbol{\epsilon}$, where $\boldsymbol{\epsilon} \sim \mathcal{N}(\mathbf{0}, \bI)$. Typically, we set $\bar{\alpha}_T \approx 0$ so that $q(\bx_T) \approx \mathcal{N}(\bx_T; \mathbf{0}, \bI)$, allowing the reverse diffusion process to start from a random Gaussian noise.

The reverse transition kernel is parameterized by the neural networks $\boldsymbol{\mu}_{\theta}$ and $\mathbf{\Sigma}_{\theta}$:
\begin{equation}
p_\theta(\bx_{t-1}|\bx_t) = \mathcal{N}\left(\bx_{t-1}; \boldsymbol{\mu}_\theta(\bx_t, t), \mathbf{\Sigma}_\theta(\bx_t, t)\right),
\label{eq:reverse}
\end{equation}
where $\theta$ denotes $p_{\theta}$'s learnable parameters. 
The goal is to maximize the likelihood of the training sample $\bx_0$ by optimizing $p_\theta(\bx_0)$. This is achieved by minimizing the variational lower bound of the negative log-likelihood $\mathbb{E}[-\log p_\theta(\bx_0)]$.

DDPM simplifies the covariance matrix $\mathbf{\Sigma}_\theta$ in \cref{eq:reverse} to a constant-scaled matrix $\tilde{\beta}_t \bI$, where $\tilde{\beta}_t=\frac{1-\bar{\alpha}_{t-1}}{1-\bar{\alpha}_t}(1-\alpha_t)$ varies across each step to control noise. Additionally, the mean $\boldsymbol{\mu}$ in \cref{eq:reverse} is expressed as a function of a learnable noise term:
\begin{equation}
\boldsymbol{\mu}_\theta(\bx_t, t) = \frac{1}{\sqrt{\alpha_t}} \left(\bx_t - \frac{1-\alpha_t}{\sqrt{1-\bar{\alpha}_t}} \boldsymbol{\epsilon}_\theta(\bx_t, t) \right),
\end{equation}
where $\boldsymbol{\epsilon}_\theta$ is a network that predicts noise $\boldsymbol{\epsilon}$ for $\bx_t$ and $t$.
According to the property in \cref{eq:forward-property} and discarding the weight, DDPM simplifies the objective function to:
\begin{equation}
\mathbb{E}_{t,\bx_0,\boldsymbol{\epsilon}} \left[\left\| \boldsymbol{\epsilon} - \boldsymbol{\epsilon}_\theta \left( \sqrt{\bar{\alpha}_t} \bx_0 + \sqrt{1 - \bar{\alpha}_t} \boldsymbol{\epsilon}, t \right) \right\|^2 \right].
\end{equation}
Eventually, samples are generated by removing noise from $\bx_T \sim \mathcal{N}(\bx_T; \mathbf{0}, \bI)$. Specifically, for $t = T, T-1, ..., 1$,
\begin{equation}
\bx_{t-1} \leftarrow \frac{1}{\sqrt{\alpha_{t}}} (\bx_{t} - \frac{1-\alpha_t}{\sqrt{1-\bar{\alpha}_t}} \boldsymbol{\epsilon}_\theta(\bx_{t}, t)) + \sigma_t \bz,
\end{equation}
where $\bz \sim \mathcal{N}(\mathbf{0}, \bI)$ for $t = T, ..., 2$, and $\bz = \mathbf{0}$ for $t = 1$.

DDPM has been widely applied in generating molecules. Considering that DDPM is designed based on continuous data space, it is more commonly used for generating continuous 3D molecular structures, such as in EDM~\citep{EDM} and GeoDiff~\citep{GeoDiff}. For discrete 2D molecular structures, the discrete version of DDPM, Discrete Denoising Diffusion Probabilistic Model (D3PM), is typically employed, as exemplified by DiGress~\citep{DiGress}.

\subsection{Score Matching with Langevin Dynamics (SMLDs)}

SMLD~\citep{SMLD}, also known as score-based generative model (SGM), uses score matching theory to learn the score function of the data distribution,
and combines it with Langevin dynamics for sampling. It comprises two main components: score matching and annealed Langevin dynamics (ALD). ALD generates samples iteratively using Langevin Monte Carlo, relying on the Stein score of a density function $q(\bx)$, defined as $\nabla_\bx \log q(\bx)$. Since the true distribution $q(\bx)$ is often unknown, score matching~\citep{ScoreMatching} approximates the Stein score with a neural network.

For efficiency, variants of score matching, such as denoising score matching~\citep{DenoisingScoreMatching},
are often used in practice.
Denoising score matching processes observed data using the forward transition kernel $q(\bx_t|\bx_0) = \mathcal{N}(\bx_t; \bx_0, \sigma_t^2 \bI)$, where $\sigma_t^2$ is a seiries of increasing noise levels for $t = 1, \ldots, T$, and then jointly estimates the Stein scores for the noise density functions $q_{\sigma_1}(\bx), q_{\sigma_2}(\bx), ..., q_{\sigma_T}(\bx)$.
The Stein score is approximated by a neural network $\bs_\theta(\bx, t)$ with $\theta$ as its learnable parameters. Thus, the objective function is defined as follows: 
\begin{equation}
\mathbb{E}_{t, \bx_0, \bx_t} \left[ \left\| \bs_\theta(\bx_t, t) - \nabla_{\bx_t} \log q_{\sigma_t}(\bx_t) \right\|^2 \right]. 
\end{equation}
With the Gaussian assumption of the forward transition kernel, the objective function can be rewritten as a tractable version: 
\begin{equation}
\mathbb{E}_{t, \bx_0, \bx_t} \left[ \delta(t) \left\| \bs_\theta(\bx_t, t) + \frac{\bx_t - \bx_0}{\sigma_t^2} \right\|^2 \right], 
\end{equation}
where $\delta(t)$ is a positive weight that depends on the noise scale $\sigma_t$. Once the score-matching network $\bs_\theta$ is trained, the ALD algorithm is used for sampling. It begins with a sequence of increasing noise levels $\sigma_1, \ldots, \sigma_T$ and an starting point \(\bx_{T,0} \sim \mathcal{N}(\mathbf{0}, \bI)\). For \(t = T, T - 1, \ldots, 0\), \(\bx_t\) is updated through \(N\) iterations that compute the following steps:
\begin{equation}
\bx_{t,n} \leftarrow \bx_{t,n-1} + \frac{1}{2} \eta_t \bs_\theta \left( \bx_{t,n-1}, t \right) + \sqrt{\eta_t} \bz, 
\end{equation}
where $\bz \sim \mathcal{N}(\mathbf{0}, \bI)$, $n = 1, \dots, N$, and $\eta_t$ is the update step. After $N$ iterations, the resulting $\bx_{t,N}$ becomes the starting point for the next $N$ iterations. $\bx_{0,N}$ will be final sample. 

SMLD has also been applied to generating molecules, such as in MDM~\citep{MDM} and LDM-3DG~\citep{LDM-3DG}. Subsequent studies demonstrat that SMLD and DDPM are theoretically equivalent and can both be regarded as discretizations of the SDE introduced in the next subsection.

\subsection{Stochastic Differential Equations (SDEs)}

Both DDPM and SMLD rely on discrete processes, requiring careful design of diffusion steps. \cite{SDE} formulate the forward process as an SDE, extending the discrete methods to continuous time space. The reverse process is modeled as a time-reverse SDE, enabling sampling by solving it. Let $\bw$ and $\bar{\bw}$ denote a standard Wiener process and its time-reverse, with continuous diffusion time $t \in [0, T]$. A general SDE is:
\begin{equation}
d\bx = f(\bx, t)dt + g(t)d\bw,
\end{equation}
where $f(\bx, t)$ and $g(t)$ are the drift coefficient and the diffusion coefficient for the diffusion process, respectively. 

The corresponding time-reverse SDE is defined as:
\begin{equation}
d\bx = \left[ f(\bx, t) - g(t)^2 \nabla_\bx \log q_t(\bx) \right] dt + g(t)d\bar{\bw}.
\end{equation}
Sampling from the probability flow ordinary differential equation (ODE) has the same distribution as the time-reverse SDE:
\begin{equation}
d\bx = \left[ f(\bx, t) - \frac{1}{2} g(t)^2 \nabla_\bx \log q_t(\bx) \right] dt.
\end{equation}
Here  $\nabla_\bx \log q_t(\bx)$ is the Stein score of the marginal
distribution of $\bx_t$, which is unknown but can be learned with a similar method as in
SMLD with the objective function:
\begin{equation}
\mathbb{E}_{t, \bx_0, \bx_t} \left[ \delta(t) \left\| \bs_\theta(\bx_t, t) - \nabla_{\bx_t} \log q_{0t}(\bx_t | \bx_0) \right\|^2 \right].
\label{eq:sde-objective}
\end{equation}

DDPM and SMLD can be regarded as discretizations of two SDEs.
Recall that $\alpha_t$ is a defined in DDPM and $\sigma_t^2$ denotes the noise level in SMLD. The SDE corresponding to DDPM is known as variance preserving (VP) SDE, defined as:
\begin{equation}
d\bx = -\frac{1}{2} \alpha(t)\bx dt + \sqrt{\alpha(t)}d\bw,
\end{equation}
where $\alpha(\cdot)$ is a continuous function, and $\alpha\left(\frac{t}{T}\right) = T(1 - \alpha_t)$ as $T \to \infty$. For the forward process of SMLD, the associated SDE is known as variance exploding (VE) SDE, defined
as:
\begin{equation}
d\bx = \sqrt{\frac{d \left[ \sigma(t)^2 \right]}{dt}} d\bw,
\end{equation}
where $\sigma(\cdot)$ is a continuous function, and $\sigma\left(\frac{t}{T}\right) = \sigma_t$ as $T \to \infty$.
Inspired by VP SDE, sub-VP SDE is designed and
performs especially well on likelihoods, given by:
\begin{equation}
d\bx = -\frac{1}{2} \alpha(t)\bx dt + \sqrt{\alpha(t) \left( 1 - e^{-2 \int_0^t \alpha(s)ds} \right)} d\bw.
\end{equation}

The objective function in \cref{eq:sde-objective} involves a perturbation distribution $q_{0t}(\bx_t | \bx_0)$ that varies for
different SDEs (i.e., VP SDE, VE SDE, sub-VP SDE). 
After $\bs_\theta(\bx, t)$ is trained, samples can be generated by solving the time-reverse SDE or the probability flow ODE with techniques such as ALD.

Because SDEs provide a continuous and flexible formulation that allows for improved control over generation processes, they have gradually replaced discrete-time formulations like DDPM and SMLD in molecular generative tasks~\citep{GDSS,EEGSDE,JODO}.

\subsection{More Variants}

These three formulations establish the theoretical foundation of diffusion models and demonstrate excellent performance in generative tasks. Building on them, diffusion models have spawned many variants and extensions aimed at enhancing generation efficiency or expanding application scenarios.
For example, 
\textit{Discrete Denoising Diffusion Probabilistic Models (D3PMs)}~\citep{D3PM} extend the DDPM to discrete data space, such as text or graphs. 
\textit{Latent Diffusion Models (LDMs)}~\citep{LDM} perform the diffusion process in latent space, significantly reducing computational complexity while maintaining generation quality.
\textit{Consistency Models (CMs)}~\citep{ConsistencyModel} focus on learning a single-step mapping from noise to data, enabling fast and high-quality sampling while maintaining consistency with the underlying data distribution.
\textit{Diffusion Bridges (DBs)}~\citep{DSB,SB-FBSDE,DDBM} extend diffusion models for generative tasks that connect different distributions, enabling efficient generation from one distribution to another.

These formulations propose innovative solutions tailored to different tasks, driving the widespread application in multi-modal generative tasks such as image, text, video, and graph~\citep{DiffusionSurvey1,DiffusionGraphSurvey1,DiffusionGraphSurvey2}.
\tikzstyle{leaf}=[draw=hiddendraw,
    rounded corners, minimum height=1em,
    fill=blue!7, text opacity=1, align=center,
    fill opacity=.5, text=black, align=left, font=\scriptsize,
    inner xsep=3pt,
    inner ysep=1pt,
    ]
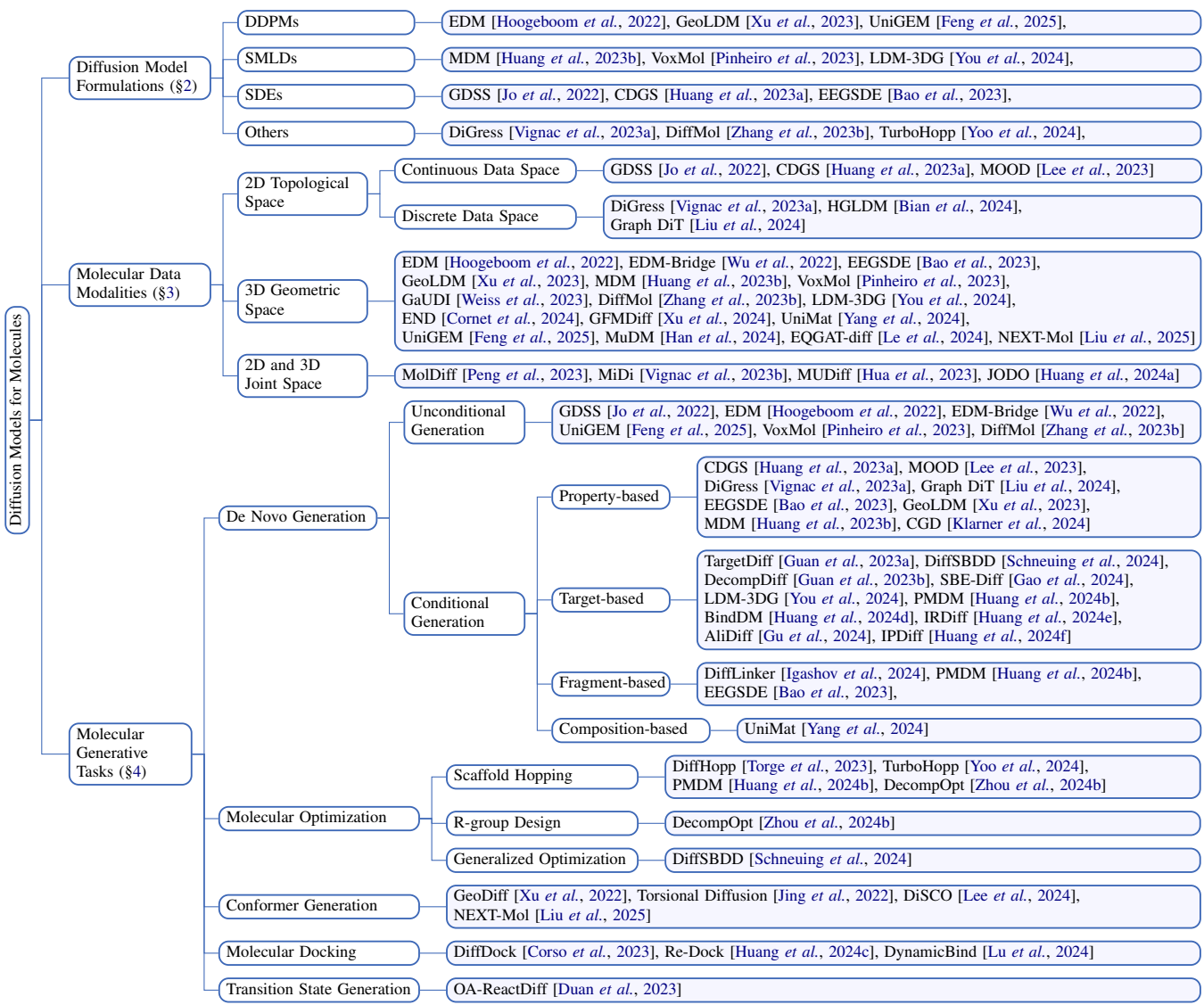
\begin{figure*}[ht]
\centering
\begin{forest}
  for tree={
  forked edges,
  grow=east,
  reversed=true,
  anchor=base west,
  parent anchor=east,
  child anchor=west,
  base=middle,
  font=\scriptsize,
  rectangle,
  draw=hiddendraw,
  rounded corners,align=left,
  line width=0.7pt,
  minimum width=2em,
    s sep=5pt,
    inner xsep=3pt,
    inner ysep=1pt,
  },
  where level=1{text width=4.5em}{},
  where level=2{text width=6em,font=\scriptsize}{},
  where level=3{font=\scriptsize}{},
  where level=4{font=\scriptsize}{},
  where level=5{font=\scriptsize}{},
  [Diffusion Models for Molecules,rotate=90,anchor=north,edge=hiddendraw
    [Diffusion Model\\Formulations (\S\ref{sec:diffusion}),edge=hiddendraw,text width=5.38em
        [DDPMs, text width=6.88em, edge=hiddendraw
            [EDM~\citep{EDM}{,}
            GeoLDM~\citep{GeoLDM}{,}
            UniGEM~\citep{UniGEM}{,},leaf,text width=31.58em, edge=hiddendraw]
        ]
        [SMLDs, text width=6.88em, edge=hiddendraw
            [MDM~\citep{MDM}{,}
            VoxMol~\citep{VoxMol}{,}
            LDM-3DG~\citep{LDM-3DG}{,},leaf,text width=31.58em, edge=hiddendraw]
        ]
        [SDEs, text width=6.88em, edge=hiddendraw
            [GDSS~\citep{GDSS}{,}
            CDGS~\citep{CDGS}{,}
            EEGSDE~\citep{EEGSDE}{,},leaf,text width=31.58em, edge=hiddendraw]
        ]
        [Others, text width=6.88em, edge=hiddendraw
            [DiGress~\citep{DiGress}{,}
            DiffMol~\citep{DiffMol}{,}
            TurboHopp~\citep{TurboHopp}{,},leaf,text width=31.58em, edge=hiddendraw]
        ]
    ]
    [Molecular Data\\Modalities (\S\ref{sec:data}),edge=hiddendraw,text width=5.38em
        [2D Topological \\Space, text width=4.88em, edge=hiddendraw
            [Continuous Data Space, edge=hiddendraw, text width=7.08em
                [GDSS~\citep{GDSS}{,} CDGS~\citep{CDGS}{,} MOOD~\citep{MOOD},leaf,text width=24.75em, edge=hiddendraw]
            ]
            [Discrete Data Space, edge=hiddendraw, text width=7.08em
                [DiGress~\citep{DiGress}{,}
                HGLDM~\citep{HGLDM}{,}\\
                Graph DiT~\citep{GraphDiT}
                ,leaf,text width=24.75em, edge=hiddendraw]
            ]
         ]
        [3D Geometric \\Space, text width=4.88em, edge=hiddendraw
            [EDM~\citep{EDM}{,}
            EDM-Bridge~\citep{EDM-Bridge}{,}
            EEGSDE~\citep{EEGSDE}{,}\\
            GeoLDM~\citep{GeoLDM}{,}
            MDM~\citep{MDM}{,}
            VoxMol~\citep{VoxMol}{,}\\
            GaUDI~\citep{GaUDI}{,}
            DiffMol~\citep{DiffMol}{,}
            LDM-3DG~\citep{LDM-3DG}{,}\\
            END~\citep{END}{,}
            GFMDiff~\citep{GFMDiff}{,}
            UniMat~\citep{UniMat}{,}\\
            UniGEM~\citep{UniGEM}{,}
            MuDM~\citep{MuDM}{,}
            EQGAT-diff~\citep{EQGAT-diff}{,}
            NEXT-Mol~\citep{NEXT-Mol}
            ,leaf,text width=33.58em, edge=hiddendraw]
        ]
        [2D and 3D \\Joint Space, text width=4.88em, edge=hiddendraw
            [MolDiff~\citep{MolDiff}{,}
            MiDi~\citep{MiDi}{,}
            MUDiff~\citep{MUDiff}{,}
            JODO~\citep{JODO}
            ,leaf,text width=33.58em, edge=hiddendraw]
        ]
    ]
    [Molecular\\Generative\\Tasks (\S\ref{sec:task}), edge=hiddendraw, text width=4.58em
        [De Novo Generation, text width=6.08em, edge=hiddendraw
            [Unconditional \\Generation,text width=4.5em, edge=hiddendraw
                [GDSS~\citep{GDSS}{,} EDM~\citep{EDM}{,} EDM-Bridge~\citep{EDM-Bridge}{,} \\
                UniGEM~\citep{UniGEM}{,} VoxMol~\citep{VoxMol}{,}
                DiffMol~\citep{DiffMol},leaf,text width=26.9em, edge=hiddendraw]
            ]
            [Conditional \\Generation, text width=4.5em, edge=hiddendraw
                [Property-based, edge=hiddendraw, text width=4.38em
                    [CDGS~\citep{CDGS}{,} MOOD~\citep{MOOD}{,}\\
                    DiGress~\citep{DiGress}{,} Graph DiT~\citep{GraphDiT}{,}\\
                    EEGSDE~\citep{EEGSDE}{,}
                    GeoLDM~\citep{GeoLDM}{,}\\
                    MDM~\citep{MDM}{,}
                    CGD~\citep{CGD},leaf,text width=20.75em, edge=hiddendraw]
                ]
                [Target-based, edge=hiddendraw, text width=4.38em
                    [TargetDiff~\citep{TargetDiff}{,}
                    DiffSBDD~\citep{DiffSBDD}{,}\\
                    DecompDiff~\citep{DecompDiff}{,}
                    SBE-Diff~\citep{SBE-Diff}{,}\\
                    LDM-3DG~\citep{LDM-3DG}{,}
                    PMDM~\citep{PMDM}{,}\\
                    BindDM~\citep{BindDM}{,}
                    IRDiff~\citep{IRDiff}{,}\\
                    AliDiff~\citep{AliDiff}{,}
                    IPDiff~\citep{IPDiff}
                    ,leaf,text width=20.75em, edge=hiddendraw]
                ]
                [Fragment-based, edge=hiddendraw, text width=4.38em
                    [DiffLinker~\citep{DiffLinker}{,} 
                    PMDM~\citep{PMDM}{,}\\
                    EEGSDE~\citep{EEGSDE}{,}
                    ,leaf,text width=20.75em, edge=hiddendraw]
                ]
                [Composition-based, edge=hiddendraw, text width=6.08em
                    [UniMat~\citep{UniMat},leaf,text width=19.05em, edge=hiddendraw]
                ]
            ]
        ]
        [Molecular Optimization,text width=7.88em, edge=hiddendraw
            [Scaffold Hopping, text width=7.5em, edge=hiddendraw
                [DiffHopp~\citep{DiffHopp}{,}
                TurboHopp~\citep{TurboHopp}{,}\\
                PMDM~\citep{PMDM}{,}
                DecompOpt~\citep{DecompOpt}
                ,leaf,text width=22.08em, edge=hiddendraw]
            ]
            [R-group Design,text width=7.5em, edge=hiddendraw
                [DecompOpt~\citep{DecompOpt},leaf,text width=22.08em, edge=hiddendraw]
            ]
            [Generalized Optimization,text width=7.5em, edge=hiddendraw
                [DiffSBDD~\citep{DiffSBDD},leaf,text width=22.08em, edge=hiddendraw]
            ]
        ]
        [Conformer Generation,text width=7.88em, edge=hiddendraw
            [GeoDiff~\citep{GeoDiff}{,}
            Torsional Diffusion~\citep{TorsionalDiffusion}{,}
            DiSCO~\citep{DiSCO}{,}\\
            NEXT-Mol~\citep{NEXT-Mol}
            ,leaf,text width=31.36em, edge=hiddendraw]
        ]
        [Molecular Docking,text width=7.88em, edge=hiddendraw
            [DiffDock~\citep{DiffDock}{,}
            Re-Dock~\citep{Re-Dock}{,}
            DynamicBind~\citep{DynamicBind}
            ,leaf,text width=31.36em, edge=hiddendraw]
        ]
        [Transition State Generation,text width=7.88em, edge=hiddendraw
            [OA-ReactDiff~\citep{OA-ReactDiff}
            ,leaf,text width=31.36em, edge=hiddendraw]
        ]
    ]
  ]
\end{forest}
\caption{A taxonomy of diffusion models for molecules with representative works.}
\label{fig:taxonomy}
\end{figure*}

\section{Molecular Data Modalities}\label{sec:data}

In molecular generative modeling, molecular data can be featurized in various modalities. The primary modalities are graphs in 2D topological space and conformations in 3D geometric space. In 2D graphs, nodes denote atoms, and edges represent bonds. In 3D conformations, the molecule is described as a point cloud, where the points are atoms with positional coordinates. 
Molecules can also be featurized in other ways, such as SMILES in 1D space and molecular descriptors. However, these featurizations are typically not generated using diffusion models and thus fall outside the scope of this survey.

Different molecular generative methods often target distinct modalities, resulting in different advantages. As illustrated in \cref{fig:taxonomy} and \cref{tab:summarizations}, existing methods are categorized into three groups based on the modality of generated molecular data: those generating molecules only in 2D space, only in 3D space, and in 2D and 3D joint space.

\subsection{Generating Molecules in 2D Topological Space}
\paratitle{Definition 1: 2D molecular graph.} The molecular graphs have categorical node and edge types, represented by the spaces $\cX$ and $\cE$, respectively, with cardinalities $a$ and $b$. A molecular graph can be denoted as $\cM_{\text{2D}} = (\bX, \bA)$, where $\bX \in \mathbb{R}^{n \times a}$ is the one-hot encoded atom feature matrix representing atom types, and $\bA \in \mathbb{R}^{n \times n \times b}$ is the one-hot encoded adjacency matrix indicating bond existence and bond types. Here, $n$ denotes the number of atoms.

The primary challenges in generating molecular graphs is maintaining permutation invariance and modeling the complex dependencies between nodes (atoms) and edges (bonds). To address these challenges, graph neural networks (GNNs) are commonly employed as the backbone architecture. As illustrated in \cref{fig:taxonomy}, several methods, such as GDSS~\citep{GDSS}, CDGS~\citep{CDGS}, and MOOD~\citep{MOOD}, directly apply diffusion models from continuous data spaces to the task of molecular graph generation. 

However, the discrete nature of atom and bond types presents further challenges. Recognizing this, DiGress~\citep{DiGress} employs a discrete diffusion model, D3PM~\citep{D3PM}, specifically designed for discrete data spaces, to model molecular graphs. On smaller molecules, discrete models like DiGress achieve results comparable to continuous models but offer the advantage of faster training.

The drawback of generating molecules in 2D space is that it only produces atom types and binding topology, lacking 3D geometric structure. Molecules inherently exist in 3D space, and their 3D structure affects their quantum properties.
This makes it unsuitable for quantum-property-based or structure-based drug design, which rely heavily on 3D structures.

In conclusion, while 2D molecular graph generation provides a useful framework for certain applications, its limitations highlight the necessity for further research into methods that can integrate 3D structural data to enhance the accuracy and applicability of molecular models in real-world scenarios.

\begin{table*}[t]
\resizebox{\linewidth}{!}{
\begin{tabular}{l|l|llllll|lll}
\toprule
& \textbf{Model} & \textbf{Diffusion Type} & \textbf{Output} & \textbf{Data Space}  & \textbf{Time Space} & \textbf{Network}  & \textbf{Objective}  & \textbf{Task} & \textbf{Condition} & \textbf{Dataset} \\ \midrule

\multirow{5}{*}{\rotatebox{90}{2D Space}}  & GDSS~\citep{GDSS} & SDE & X, A & Cont. & Cont. & MPNN & score matching & De Novo Gen. & None & ZINC250k, QM9-2014 \\
 & CDGS~\citep{CDGS} & SDE & X, A & Cont. & Cont. & MPNN+GT & noise prediction & De Novo Gen. & None/Property & ZINC250k, QM9-2014 \\
 & MOOD~\citep{MOOD} & SDE & X, A & Cont. & Cont. & MPNN & score matching & De Novo Gen. & OOD-ness/Property & ZINC250k \\
 & DiGress~\citep{DiGress} & D3PM & X, A & Disc. & Disc. & MPNN & data prediction & De Novo Gen. & None/Property & QM9-2014, MOSES, GuacaMol \\
 & Graph DiT~\citep{GraphDiT} & D3PM & X, A & Disc. & Disc. & GT & data prediction & De Novo Gen. & Property & Polymer, MoleculeNet \\
\midrule
\multirow{25}{*}{\rotatebox{90}{3D Space}} & EDM~\citep{EDM} & DDPM & X, P & Cont. & Disc. & EGNN & noise prediction & De Novo Gen. & None/Property & QM9-2014, GEOM-Drugs \\
 & EDM-Bridge~\citep{EDM-Bridge} & DB & X, P & Cont. & Cont. & EGNN & score matching & De Novo Gen. & None & QM9-2014, GEOM-Drug \\
 & EEGSDE~\citep{EEGSDE} & SDE & X, P & Cont. & Cont. & EGNN & noise prediction & De Novo Gen. & Property/Substructure & QM9-2014, GEOM-Drug \\
 & GeoLDM~\citep{GeoLDM} & DDPM & Latent & Cont. & Disc. & EGNN & noise prediction & De Novo Gen. & None/Property & QM9-2014, GEOM-Drug \\
 & MDM~\citep{MDM} & SMLD & X, P & Cont. & Disc. & EGNN & score matching & De Novo Gen. & None/Property & QM9-2014, GEOM-Drug \\
 & UniGEM~\citep{UniGEM} & DDPM & X, P & Cont. & Disc. & EGNN & noise prediction & De Novo Gen. & None & QM9-2014, GEOM-Drug \\
 & VoxMol~\citep{VoxMol} & SMLD & VM* & Cont. & Disc. & 3D U-Net & score matching & De Novo Gen. & None & QM9-2014, GEOM-Drug \\
 & DiffMol~\citep{DiffMol} & D3PM & X, P & Disc. & Disc. & EGNN & \makecell[l]{score matching\\ \&data prediction} & De Novo Gen. & None & QM9-2014 \\
 & LDM-3DG~\citep{LDM-3DG} & SMLD & Latent & Cont. & Disc. & 3D MPNN & score matching & De Novo Gen. & Property/Target & \makecell[l]{QM9-2014, GEOM-Drug,\\ CrossDocked} \\
 & GaUDI~\citep{GaUDI} & DDPM & GoR* & Cont. & Disc. & EGNN & noise prediction & De Novo Gen. & Property & COMPAS-1x, PAS \\
 & GeoDiff~\citep{GeoDiff} & DDPM & P & Cont. & Disc. & GFN & noise prediction & Conformer Gen. & Molecular Graph & GEOM-QM9, GEOM-Drugs \\
 & Torsional Diffusion~\citep{TorsionalDiffusion} & SDE & P & Cont. & Cont. & GFN & score matching & Conformer Gen. & Molecular Graph & \makecell[l]{GEOM-QM9, GEOM-Drugs,\\ GEOM-XL} \\
 & DiffLinker~\citep{DiffLinker} & DDPM & Linker & Cont. & Disc. & EGNN & noise prediction & De Novo Gen. & 3D Fragments & ZINC, CASF, GEOM \\
 & OA-ReactDiff~\citep{OA-ReactDiff} & DDPM & TS* & Cont. & Disc. & LEFTNet & noise prediction & TS Gen. & Reactant and Product & Transition1x \\
 & TargetDiff~\citep{TargetDiff} & DDPM+D3PM & X, P & Cont.+Disc. & Disc. & EGNN & \makecell[l]{data prediction} & De Novo Gen. & Target & CrossDocked \\
 & DiffSBDD~\citep{DiffSBDD} & DDPM & X, P & Cont. & Disc. & EGNN & noise prediction & De Novo Gen., Opti. & Target & CrossDocked \\
 & SBE-Diff~\citep{SBE-Diff} & DDPM+D3PM & X, P & Cont.+Disc. & Disc. & EGNN & \makecell[l]{data prediction} & De Novo Gen. & Target & CrossDocked \\
 & PMDM~\citep{PMDM} & DDPM & X, P & Cont. & Disc. & SchNet & noise prediction & De Novo Gen., Opti. & Target & CrossDocked \\
 & BindDM~\citep{BindDM} & DDPM+D3PM & X, P & Cont.+Disc. & Disc. & EGNN & \makecell[l]{data prediction} & De Novo Gen. & Target & CrossDocked \\
 & DiffDock~\citep{DiffDock} & SDE & P* & Cont. & Cont. & EGNN & score matching & Docking & Protein and Ligand & PDBBind \\
 & Re-Dock~\citep{DiffDock} & DB & P* & Cont. & Cont. & EGNN & score matching & Docking & Protein and Ligand & PDBBind \\
 & DiffHopp~\citep{DiffHopp} & DDPM & X, P & Cont. & Disc. & GVP & noise prediction & Opti. & Protein-ligand Complex & PDBBind \\
 & TurboHopp~\citep{TurboHopp} & CM & X, P & Cont. & Cont. & GVP & self-consistency & Opti. & Protein-ligand Complex & PDBBind \\
 & UniMat~\citep{UniMat} & DDPM & PT* & Cont. & Disc. & Conv, Attn & noise prediction & De Novo Gen. & Composition & Perov-5, Carbon-24, MP-20 \\
\midrule
\multirow{6}{*}{\rotatebox{90}{2D\&3D Space}} & MolDiff~\citep{MolDiff} & DDPM+D3PM & X, P, A & Cont.+Disc. & Disc. & EGNN & \makecell[l]{data prediction} & De Novo Gen. & None & QM9-2014, GEOM-Drug \\
 & MiDi~\citep{MiDi} & DDPM+D3PM & X, P, A & Cont.+Disc. & Disc. & rEGNN & \makecell[l]{data prediction} & De Novo Gen. & None & QM9-2014, GEOM-Drug \\
 & DecompDiff~\citep{DecompDiff} & DDPM+D3PM & X, P, A & Cont.+Disc. & Disc. & EGNN & \makecell[l]{data prediction} & De Novo Gen. & Target & CrossDocked \\
 & DecompOpt~\citep{DecompOpt} & DDPM+D3PM & X, P, A & Cont.+Disc. & Disc. & EGNN & \makecell[l]{data prediction} & Opti. & Target and Ligand & CrossDocked \\
 & JODO~\citep{JODO} & SDE & X, P, A & Cont. & Cont. & DGT & data prediction & De Novo Gen. & None/Property & QM9-2014, GEOM-Drug \\
 & MUDiff~\citep{MUDiff} & DDPM+D3PM & X, P, A & Cont.+Disc. & Disc. & MUformer & \makecell[l]{noise\&data prediction} & De Novo Gen. & None/Property & QM9-2014 \\
 \bottomrule
\end{tabular}}
\caption{A comprehensive summary of diffusion models for molecules in the literature, categorized based on the data modality into three groups: diffusion in 2D space, in 3D space, and in joint space.
Acronyms in \textbf{Data Space} and \textbf{Time Space}: Cont. refers to continuous space; Disc. refers to discrete space.
Acronyms in \textbf{Task}: Gen. refers to generation task; Opti. refers to optimization task.
Acronyms in \textbf{Output}: X, A, and P refer to atom features, adjacency matrix, and positions, respectively; VM denotes voxelized molecules; GoR refers to graph of rings; TS represents transition states, P* refers to ligand poses in the submanifold space; PT refers to periodic table-based molecular representations.
}
\label{tab:summarizations}
\end{table*}

\subsection{Generating Molecules in 3D Geometric Space}
\paratitle{Definition 2: 3D molecular conformation.} A 3D molecular conformation is formally defined as $\cM_{\text{3D}} = (\bX, \bP)$, where $\bX \in \mathbb{R}^{n \times a}$ is the one-hot encoded atom feature matrix representing atom types, $\bP \in \mathbb{R}^{n \times 3}$ is the position coordinate matrix of atoms, and $n$ denotes the number of atoms. 

Methods for generating molecules in 3D space focus on producing both the atom types and their geometric structures, while not directly generating the binding topology. The binding topology can be inferred through post-processing steps. A significant challenge in these methods is maintaining SE(3) equivariance, which ensures that the generated molecular structures are invariant to transformations such as rotations and translations in 3D space. This is often achieved by integrating equivariant graph neural networks~\citep{EGNN} and employing techniques like zero center of mass (CoM) adjustments~\citep{EDM}.

The drawback of these methods is that they do not consider binding topology during the generation of geometric structures. Binding information is crucial for evaluating the quality of molecular generation and many downstream tasks. Post-processing to infer binding topology can introduce errors and often results in suboptimal solutions.
Moreover, for larger molecules, such as those in GEOM-Drug~\citep{GEOM-datasets}, directly generating high-quality 3D structures is challenging. Incorporating 2D topology during the generation process can provide valuable guidance and improve the quality of the resulting molecular structures.

\subsection{Generating Molecules in 2D\&3D Joint Space}

The generation of molecular structures in 2D and 3D joint spaces offers a comprehensive approach to capturing both the topological and geometric properties of molecules. This dual featurization, also referred to as a complete molecular structure in some works~\citep{JODO,MUDiff}, is crucial for accurately depicting molecules.

\paratitle{Definition 3: Complete molecular structure.} A molecule in this joint space is defined as $\cM = (\bX, \bA, \bP)$, where $\bX \in \mathbb{R}^{n \times a}$ is the atom feature matrix, $\bA \in \mathbb{R}^{n \times n \times b}$ is the one-hot encoded adjacency matrix indicating bond existence and bond types, and $\bP \in \mathbb{R}^{n \times 3}$ is the position coordinate matrix of the atoms. Here, $n$ denotes the number of atoms in the molecule.

Methods that generate molecules in both 2D and 3D spaces simultaneously focus on creating a cohesive representation where the 2D topological structure and the 3D geometric structure interact and complement each other during the generation process. 
The 2D representation captures the bonding relationships between atoms, providing essential information about the molecular connectivity and chemical structure.
The 3D representation provides spatial information, crucial for understanding the molecule's shape, conformation, and potential interactions with other molecules.
By integrating these two aspects, the generation process benefits from the strengths of both representations. The 2D topology can guide the 3D structure formation, ensuring that spatial arrangements are chemically feasible, while the 3D geometry can refine the 2D topology by suggesting plausible bonding patterns based on spatial proximity~\citep{MolDiff,MiDi}.
The interaction allows for continuous refinement and correction, leading to more stable and realistic molecular models.

The main technical challenge in generating molecules in joint space is managing the discrete topological structures alongside continuous geometric structures. JODO~\citep{JODO} uses SDE to treat both structures as continuous variables. Conversely, MUDiff~\citep{MUDiff} employs discrete D3PM for topology and continuous DDPM for geometry, handling them separately. The elegant and simultaneous generation of molecular structures across these modalities remains a significant area for exploration.
\section{Molecular Generative Tasks}\label{sec:task}

In this section, we explore various molecular generative tasks that leverage diffusion models, as outlined in \cref{fig:taxonomy}. These tasks are crucial for advancing molecular design and discovery, providing innovative solutions across different domains.

\subsection{De Novo Generation}
De novo generation involves creating novel molecular structures from scratch. This approach is essential for discovering new compounds without relying on existing molecular templates. It includes two main sub-tasks: unconditional and conditional generation.

\paratitle{Unconditional generation.}
Unconditional generation focuses on producing molecules without specific constraints. Starting from a random noise vector, these models generate entirely new molecular structures, exploring the vast chemical space for novel applications.

\paratitle{Conditional Generation}
Conditional generation tailors molecule creation based on specific conditions, such as desired properties, targets, or fragments, allowing for more directed and efficient molecular design. By incorporating these conditions, diffusion models produce molecules that meet predefined criteria. 
Existing works can be further divided into four categories based on the type of conditions applied.

\textit{Property-based molecular generation}, also known as \textit{inverse molecule design}, aims to generate molecules with desired properties such as bioactivity and synthesizability.
More specifically, the inverse molecule design can be further divided into single-property conditioning~\citep{CDGS,EDM,GeoLDM,MDM} and multiple-property conditioning~\citep{DiGress,GraphDiT,EEGSDE}.
Among them, MOOD~\citep{MOOD} and CGD~\citep{CGD} also focus on generating structurally novel molecules outside the training distribution, referred to as OOD molecule generation.

\textit{Target-based molecular generation}, also known as \textit{structure-based drug design (SBDD)}, generates molecules based on the 3D structure of target binding pockets, aiming to enhance interaction with specific targets.
IRDiff~\citep{IRDiff} introduces interaction-based retrieval to generate target-specific molecules based on retrieved high-affinity ligand references.

\textit{Fragment-based molecular generation} specifies the generation of molecules with particular fragments. DiffLinker~\citep{DiffLinker} focuses on linker design, generating linkers that connect fragments into a complete molecule.

\textit{Composition-based molecular generation} restricts the elemental composition of generated molecules, ensuring they meet specific compositional criteria~\citep{UniMat}.

\subsection{Molecular Optimization}
Molecular optimization tasks aim to improve existing molecules for better performance or properties, differing from de novo generation by focusing on modifying known structures rather than creating new ones. Starting with an existing molecule, these tasks refine it to enhance its properties or performance. This task includes scaffold hopping, R-group design, and generalized optimization.

\textit{Scaffold hopping} involves modifying molecular scaffolds to discover new compounds, transforming known scaffolds into new ones that retain biological activity~\citep{DiffHopp}.

\textit{R-group design} focuses on optimizing molecules by fine-tuning the properties of lead compounds through the adjustment of specific R-groups~\citep{DecompOpt}.

\textit{Generalized optimization} is a flexible approach to optimize molecules without being restricted to specific strategies like scaffold hopping or R-group design. 
DiffSBDD~\citep{DiffSBDD} allows for a broader range of structural changes, as long as the optimized molecule maintains a certain level of similarity to the original structure. This flexibility enables the exploration of diverse pathways to improve properties.

\subsection{Conformer Generation}
Conformer generation involves predicting the 3D conformers of a molecule based on its 2D topological structure. This task is crucial for understanding the spatial arrangement of atoms within a molecule, which is essential for predicting molecular interactions, reactivity, and properties.
Generated 3D conformers reflect the molecule's potential energy landscape and geometric constraints.
GeoDiff~\citep{GeoDiff} and Torsional Diffusion~\citep{TorsionalDiffusion} employ diffusion models to generate 3D molecular conformers in Cartesian space and torsion angle space, respectively. DiSCO~\citep{DiSCO} further optimizes the predicted conformers with Diffusion Bridge.

\subsection{Molecular Docking}
Molecular docking tasks involve predicting how molecules interact with biological targets, a key step in drug discovery for assessing binding affinity and specificity. By analyzing a molecule and a target structure, DiffDock~\citep{DiffDock} predicts the binding pose with the diffusion model.
Re-Dock~\citep{Re-Dock} further utilizes the diffusion bridge for flexible and realistic molecular docking, which predicts the binding poses of ligands and pocket sidechains simultaneously.

\subsection{Transition State Generation}
Transition state generation focuses on predicting the 3D structure of transition states in chemical reactions, using the reactants and products as inputs. This task is vital for understanding reaction mechanisms, as the transition state represents the highest energy point along the reaction pathway. Accurate modeling of these states provides insights into reaction kinetics and can aid in the design of catalysts and optimization of reaction conditions~\citep{OA-ReactDiff}.
\section{Discussion and Future Direction}
In this section, we discuss the current state and challenges in diffusion models for molecules and outline several promising directions for future research to advance this area.

\paratitle{Complete data modality.}
Most existing works fall under the category of generating molecules in 3D space,
neglecting 2D topology.
Considering the complementary nature of 2D and 3D structures, generating molecules in a joint 2D and 3D space holds significant potential for producing more realistic molecules.
This approach has proven effective in de novo generation~\citep{MUDiff,JODO}, but its broader potential in other generative tasks remains underexplored.

\paratitle{Sophisticated diffusion models.}
As summarized in \cref{tab:summarizations}, 
the diffusion models employed in existing works exhibit a wide variety of formulations. Regarding the time space, there is a shift from discrete-time methods to more generalized continuous-time SDEs. In terms of the data space, an open challenge lies in handling the discrete molecular components (e.g., atom and bond types) alongside the continuous components (e.g., coordinates). Moreover, advanced formulations and techniques, such as flow matching and efficient sampling, remain underutilized.

\paratitle{Challenging generative tasks.}
Many existing works focus on fundamental tasks,
like unconditional or single-conditional generation, 
with insufficient attention to more practical generative tasks, such as multi-conditional generation, molecular optimization, and docking. 
Furthermore, poor performance on large molecules in GEOM-Drugs compared to small molecules in QM9, highlights room for improvement. Additionally, extending molecular generation to complex~\citep{DynamicBind} while considering inter-molecular interactions, presents a another promising yet challenging direction.

\paratitle{Expressive network architectures.}
Existing methods rely on relatively classical network architectures like EGNNs. 
Recent advances in more expressive equivariant neural networks offer new opportunities. Incorporating more powerful architectures into molecular diffusion models could further enhance their performance and effectiveness.

\paratitle{Relationship between molecular generation and molecular representation.}
With the increasing recognition of diffusion models' ability to learn representations, exploring the relationship between molecular generation and molecular representation based on diffusion models emerges as a promising direction. MoleculeSDE~\citep{MoleculeSDE}, SubgDiff~\citep{SubgDiff}, and UniGEM~\citep{UniGEM} mark pioneering steps, but there remains significant room for further research.

\section{Conclusion}

In this survey, we provide an systematic overview of diffusion model-based molecular generative methods, addressing the growing need for clarity in this rapidly evolving area. By introducing a novel taxonomy that categorizes research based on method formulations, data modalities, and task types, we offer a structured framework to navigate the research landscape. Furthermore, we identify opportunities in the existing literature, highlighting promising directions for future exploration.

\bibliographystyle{named}
\bibliography{ijcai25}

\end{document}